\documentclass{article}

\usepackage{microtype}
\usepackage{graphicx}
\usepackage{subcaption}
\usepackage{booktabs}
\usepackage{hyperref}
\usepackage{xcolor}

\usepackage[preprint]{icml2026}

\usepackage{amsmath}
\usepackage{amssymb}
\usepackage{mathtools}
\usepackage{amsthm}
\usepackage{tabularx}
\usepackage[capitalize,noabbrev]{cleveref}
\usepackage[textsize=tiny]{todonotes}
\usepackage{natbib}

\theoremstyle{plain}

\theoremstyle{definition}

\theoremstyle{remark}

\icmltitlerunning{Centrality-Based Pruning for Efficient Echo State Networks}

\begin{document}

\twocolumn[
\icmltitle{Centrality-Based Pruning for Efficient Echo State Networks}

\begin{icmlauthorlist}
\icmlauthor{Sudip Laudari}{aff}
%\icmlauthor{Dohee Jung}{aff}
%\icmlauthor{Sang-Gu Lee}{aff}
\end{icmlauthorlist}

\icmlaffiliation{aff}{Independent Researcher}

\vskip 0.05in
\icmlcorrespondingauthor{Sudip Laudari}{sudiplaudari@gmail.com}
\vskip 0.05in
]

\printAffiliationsAndNotice{Preprint}
\vspace{-0.15in}

%==============================================%
%==============================================%

\begin{abstract}
Echo State Networks (ESNs) are a reservoir computing framework widely used for nonlinear time-series prediction. However, despite their effectiveness, randomly initialized reservoirs often contain redundant nodes, leading to unnecessary computational overhead and reduced efficiency. 
In this work, we propose a graph centrality-based pruning approach that interprets the reservoir as a weighted directed graph and removes structurally less important nodes using centrality measures.
Experiments on Mackey-Glass time-series prediction and electric load forecasting demonstrate that the proposed method can significantly reduce reservoir size while maintaining, and in some cases improving, prediction accuracy.
\end{abstract}

\icmlkeywords{Echo State Networks, Reservoir Computing, Graph Centrality, Network Pruning, Time-Series Prediction}

%==============================================%
%==============================================%
\section{Introduction}

Echo State Networks (ESNs), introduced by Jaeger \citep{jaeger2001short,jaeger2007echo}, are a reservoir computing framework designed for modeling nonlinear temporal dynamics. More broadly, reservoir computing provides an efficient alternative to training recurrent neural networks, as it avoids costly optimization of recurrent connections while still achieving strong performance in time-series forecasting and chaotic systems \citep{lukovsevivcius2012reservoir}.

A key idea behind ESNs is to use a fixed, randomly initialized recurrent network, known as the reservoir \citep{maass2002real}, to transform input signals into high-dimensional dynamic representations. In this setting, only the output layer is trained, which simplifies the learning process and reduces computational cost, while still capturing complex temporal dependencies \citep{jaeger2001short}.

Due to these advantages, ESNs have been widely applied to tasks such as chaotic time-series prediction, energy load forecasting, signal processing, and control systems \citep{jaeger2004harnessing, bianchi2016investigating, pathak2018model}. They are particularly effective for modeling systems with nonlinear and long-term temporal behavior.

However, despite their efficiency, ESNs rely on randomly constructed reservoirs, which often contain redundant or weakly contributing nodes \citep{rodan2010minimum}. This can lead to unnecessary complexity and increased computational cost without a corresponding improvement in predictive performance \citep{gallicchio2018design}. As a result, several studies have explored reservoir optimization through sparsification, structural adaptation, and topology design \citep{bala2018applications, scardapane2014effective, liu2022broad}.

At the same time, neural networks can be viewed as graph-structured systems, where neurons correspond to nodes and connections correspond to edges \citep{battaglia2018relational}. This perspective allows the use of tools from graph theory to better understand network structure and information flow \citep{estrada2012structure}. In particular, centrality measures provide a principled way to quantify the relative importance of nodes in such graph-structured representations, and have been explored in the context of pruning and influence estimation in neural networks \citep{li2019deep, ayoub2022link, fan2019learning, freeman1978centrality, valente2008correlated}.

Motivated by these insights, we explore a centrality-based approach for pruning ESN reservoirs. Specifically, we model the reservoir as a weighted directed graph and use centrality measures to identify structurally less important nodes. These nodes are then removed to obtain a more compact reservoir while aiming to preserve the essential dynamics of the system. We investigate this approach on both synthetic and real-world time-series tasks, including Mackey-Glass prediction and electric load forecasting.

%============================================
\section{Related Work}

Echo State Networks (ESNs) and reservoir computing have been extensively studied for nonlinear time-series modeling, with applications in chaotic systems, energy forecasting, and signal processing \citep{jaeger2004harnessing, lukovsevivcius2012reservoir, bianchi2016investigating, pathak2018model}. Recent work has highlighted that reservoir structure, including connectivity patterns and weight distributions, plays a crucial role in determining prediction accuracy and stability, motivating research on systematic reservoir design and optimization \citep{liu2022broad, gonbadi2025input, sima2025enhancing}.

Several studies focus on improving ESN efficiency through reservoir optimization and pruning. Early work proposed minimal and deterministic reservoirs to reduce redundancy while preserving dynamical richness \citep{rodan2010minimum, ozturk2007analysis, chen2016echo}. More recent approaches introduce pruning and adaptive selection mechanisms to identify and remove less informative reservoir nodes. For example, Broad ESNs with pruning remove redundant nodes while maintaining or improving performance on air quality and power load datasets \citep{liu2022broad}, while multi-reservoir ESNs with top-down pruning strategies design sparse and structured reservoirs to improve both accuracy and training efficiency \citep{wang2024multi}. Other approaches employ correlation analysis, partial-correlation-based pruning, and adaptive selection strategies to retain only the most informative reservoir units \citep{sima2025enhancing, huang2023semi}.

In parallel, input-driven and data-dependent reservoir design has gained attention. Input-aware optimization methods suggest that reservoir topology and weight scaling should be adapted to the statistical properties of the input rather than chosen randomly \citep{gonbadi2025input}. Similarly, reservoir state selection methods based on particle swarm optimization and Bayesian optimization demonstrate that selecting salient states can significantly improve forecasting performance \citep{sima2025enhancing}. These works complement earlier studies on deep, modular, and physically inspired reservoirs, which emphasize the importance of structured architectures for capturing multiscale temporal dynamics \citep{gallicchio2017deep, sun2022systematic, nakajima2021reservoir, tanaka2019recent}.

Neural networks can also be analyzed as graph-structured systems, where neurons correspond to nodes and connections correspond to edges \citep{battaglia2018relational, estrada2012structure}. This perspective enables the use of graph theory and centrality measures to analyze node importance and information flow \citep{freeman1978centrality, estrada2005subgraph, newman2018networks}. In deep learning, centrality-based pruning methods model neural networks as graphs and use measures such as eigenvector centrality to rank and remove less important neurons, resulting in more compact models with competitive performance \citep{li2019deep, fan2019learning, chowdhury2025effective}. 

However, most existing centrality-based pruning approaches focus on feedforward or convolutional architectures. The application of graph centrality to ESN reservoirs, which are governed by recurrent dynamics, remains limited. Building on prior work in ESN pruning and reservoir optimization \citep{rodan2010minimum, scardapane2014effective, bala2018applications, liu2022broad, wang2024multi}, we propose a centrality-based pruning approach that models the ESN reservoir as a weighted directed graph and removes structurally less important nodes while preserving its essential temporal dynamics.

%=========================================
%=========================================
\section{ESN Architecture}
An Echo State Network consists of an input layer, a recurrent reservoir, and a trainable output layer. The input layer projects external signals into the reservoir, whose fixed recurrent connections generate high-dimensional dynamic states. The output layer then maps these states to predictions using trained readout weights.

\begin{figure}[!ht]
\centering
\includegraphics[width=\columnwidth, height=0.6\columnwidth]{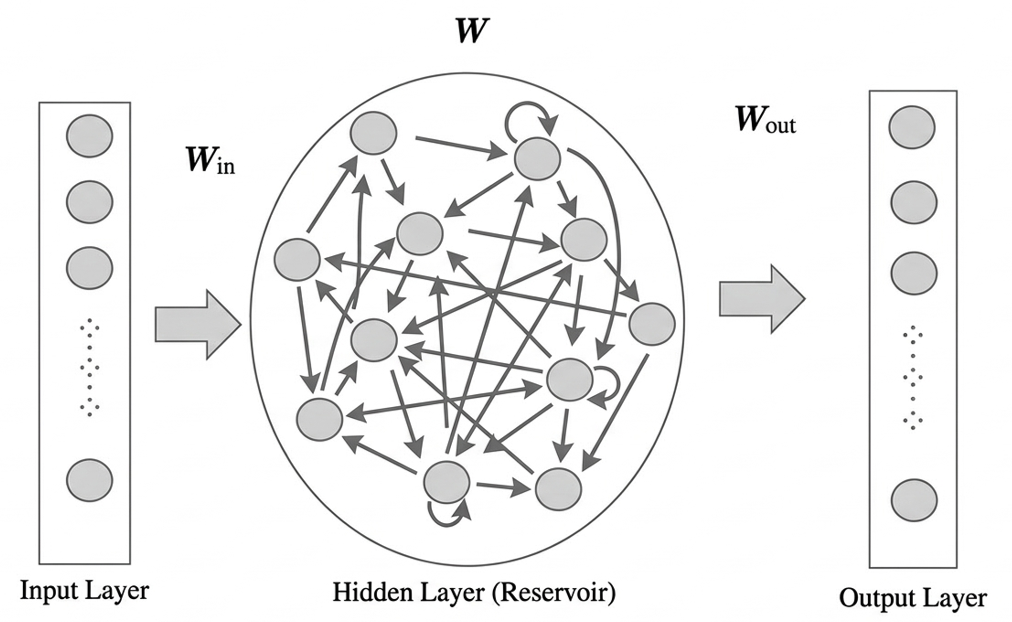}
\caption{Echo State Network (ESN) architecture. The input is mapped into a recurrent reservoir through $W^{in}$, where the internal connections $W$ generate dynamic states. The output is computed using trained weights $W^{out}$, while the reservoir remains fixed after initialization.}
\label{fig:esn}
\end{figure}

The reservoir state is updated as
\begin{equation}
x(n+1)=f\big(Wx(n)+W^{in}u(n+1)+W^{back}y(n)\big),
\end{equation}
and the output is computed as
\begin{equation}
y(n+1)=f^{out}\big(W^{out}[u(n+1),x(n+1),y(n)]\big).
\end{equation}

Here, $x(n)$, $u(n)$, and $y(n)$ denote the reservoir state, input, and output at time step $n$, respectively. The matrices $W$, $W^{in}$, and $W^{back}$ represent the reservoir, input, and feedback connections. Moreover, $W^{out}$ denotes the output weight matrix, which is the only component trained during learning.

A key property of ESNs is the echo state property, which ensures that the influence of initial states gradually vanishes over time. In practice, this is typically enforced by controlling the spectral radius of the reservoir weight matrix to maintain stable dynamics.

%==============================================

\section{Centrality Models}

To address the redundancy in ESN reservoirs, we model the reservoir as a directed weighted graph, where each neuron corresponds to a node and each connection corresponds to a weighted edge. From this perspective, the reservoir can be analyzed using tools from graph theory, where centrality measures provide a principled way to quantify the importance of individual nodes \citep{freeman1978centrality,estrada2005subgraph,bloch2023centrality,newman2018networks}. Intuitively, centrality measures capture how influential a node is within the network structure. Nodes that play a more significant role in information flow or connectivity are assigned higher centrality values, while less important nodes receive lower scores; previous work has shown that such measures correlate with influence, diffusion, and control in complex networks \citep{kitsak2010identification,valente2008correlated}.

In ESNs, reservoir states evolve through recurrent interactions, and nodes with stronger or more connected positions are expected to have a greater influence on the system dynamics. We begin with standard degree-based centrality measures. The in-degree and out-degree centralities are defined as
\begin{equation}
C_{in}(i)=|I_i|, \qquad
C_{out}(i)=|O_i|,
\end{equation}
where $|I_i|$ and $|O_i|$ denote the total incoming and outgoing connection strengths of node $i$, respectively.

However, since ESN reservoirs contain both positive and negative weights, it is important to account for their different effects. To this end, we separate positive and negative contributions. Let $|I_i^{+}|$ and $|I_i^{-}|$ denote the sums of positive and absolute negative incoming weights, respectively. Similarly, $|O_i^{+}|$ and $|O_i^{-}|$ represent the positive and absolute negative outgoing connection strengths. Since positive and negative weights contribute differently to the reservoir dynamics, separating them provides a more accurate characterization of node influence.

Based on this formulation, we define three centrality measures. The first measure is given by
\begin{equation}
C_1(i)=
\frac{|I_i^{+}|-|I_i^{-}|}
{|I_i^{+}|+|I_i^{-}|},
\end{equation}
which captures the balance between positive and negative incoming influences.

The second measure is defined as
\begin{equation}
C_2(i)=
\frac{|I_i^{+}|+|O_i^{+}|-|I_i^{-}|-|O_i^{-}|}
{|I_i^{+}|+|O_i^{+}|+|I_i^{-}|+|O_i^{-}|},
\end{equation}
which extends this idea by incorporating both incoming and outgoing connections, providing a more global view of node influence within the reservoir.

Finally, the total connectivity strength is defined as
\begin{equation}
C_3(i)=|I_i^{+}|+|O_i^{+}|+|I_i^{-}|+|O_i^{-}|.
\end{equation}
This measure reflects the overall magnitude of connectivity for each node, regardless of sign.

These centrality measures are used to rank reservoir nodes according to their structural importance. Nodes with low centrality values are considered less influential and are progressively removed, after which the output weights are retrained using the reduced reservoir. This follows centrality-based pruning strategies that remove low-importance units while maintaining performance in neural networks \citep{li2019deep,chowdhury2025effective}. 

Prior work further shows that spectral and eigenvector-based centrality measures can capture influence propagation in complex networks \citep{estrada2005subgraph,perra2008spectral,kitsak2010identification}. This supports the use of centrality as a structural criterion for pruning ESN reservoirs.

%=====================================

\begin{table*}[!t]
\centering
\caption{Reservoir size and testing error after pruning based on $C_{in}$, $C_{out}$, $C_1$, $C_2$, and $C_3$ for Mackey-Glass time-series prediction.}
\label{tab:mackey_all_models}
\begin{tabular*}{\textwidth}{@{\extracolsep{\fill}}ccccc@{}}
\toprule
Methods & Initial $N$ (RMSE) & Optimal $N$ (RMSE) & Error Reduction & Smallest $N$ \\
\midrule

$C_{in}$ 
& $N=200$ (0.010477) & $N=185$ (0.009725) & 0.000752 (7.2\%) & $N=138$ \\
& $N=300$ (0.010619) & $N=251$ (0.009734) & 0.000885 (8.3\%) & $N=202$ \\
& $N=500$ (0.009047) & $N=490$ (0.008542) & 0.000505 (5.6\%) & $N=366$ \\
& $N=700$ (0.009475) & $N=615$ (0.009082) & 0.000393 (4.1\%) & $N=458$ \\
\midrule

$C_{out}$ 
& $N=200$ (0.010477) & $N=194$ (0.009714) & 0.000763 (7.3\%) & $N=147$ \\
& $N=300$ (0.010619) & $N=294$ (0.010541) & 0.000078 (0.7\%) & $N=207$ \\
& $N=500$ (0.009047) & $N=491$ (0.008399) & 0.000648 (7.2\%) & $N=360$ \\
& $N=700$ (0.009475) & $N=671$ (0.008879) & 0.000596 (6.3\%) & $N=443$ \\
\midrule

$C_1$ 
& $N=200$ (0.010477) & $N=184$ (0.010032) & 0.000445 (4.2\%) & $N=142$ \\
& $N=300$ (0.010619) & $N=279$ (0.009972) & 0.000647 (6.1\%) & $N=196$ \\
& $N=500$ (0.009047) & $N=488$ (0.008417) & 0.000630 (7.0\%) & $N=364$ \\
& $N=700$ (0.009475) & $N=666$ (0.009085) & 0.000390 (4.1\%) & $N=444$ \\
\midrule

$C_2$ 
& $N=200$ (0.010477) & $N=182$ (0.009289) & 0.001188 (11.3\%) & $N=143$ \\
& $N=300$ (0.010619) & $N=279$ (0.009447) & 0.001172 (11.0\%) & $N=185$ \\
& $N=500$ (0.009047) & $N=470$ (0.008553) & 0.000494 (5.5\%) & $N=354$ \\
& $N=700$ (0.009475) & $N=646$ (0.008962) & 0.000513 (5.4\%) & $N=478$ \\
\midrule

$C_3$ 
& $N=200$ (0.010477) & $N=145$ (0.010363) & 0.000114 (1.1\%) & $N=145$ \\
& $N=300$ (0.010619) & $N=300$ (0.010619) & -- & $N=197$ \\
& $N=500$ (0.009047) & $N=496$ (0.008715) & 0.000332 (3.7\%) & $N=366$ \\
& $N=700$ (0.009475) & $N=628$ (0.008829) & 0.000646 (6.8\%) & $N=398$ \\
\bottomrule
\end{tabular*}
\end{table*}

\begin{table*}[!h]
\centering
\caption{Reservoir size and testing error after pruning based on $C_{in}$, $C_{out}$, $C_1$, $C_2$, and $C_3$ for electricity load forecasting.}
\label{tab:electricity_all_models}
\begin{tabular*}{\textwidth}{@{\extracolsep{\fill}}ccccc@{}}
\toprule
Methods & Initial $N$ (RMSE) & Optimal $N$ (RMSE) & Error Reduction & Smallest $N$ \\
\midrule

$C_{in}$ 
& $N=200$ (0.244586) & $N=195$ (0.149177) & 0.095409 (39\%) & 168 \\
& $N=300$ (0.358582) & $N=300$ (0.358582) & -- & 284 \\
& $N=500$ (0.405852) & $N=485$ (0.355074) & 0.050778 (12.5\%) & 468 \\
& $N=700$ (0.635077) & $N=684$ (0.607292) & 0.027785 (4.4\%) & 674 \\
\midrule

$C_{out}$ 
& $N=200$ (0.244586) & $N=199$ (0.157439) & 0.087147 (35.6\%) & 192 \\
& $N=300$ (0.358582) & $N=298$ (0.309402) & 0.049180 (13.7\%) & 285 \\
& $N=500$ (0.405852) & $N=463$ (0.224701) & 0.181151 (44.6\%) & 402 \\
& $N=700$ (0.635077) & $N=637$ (0.514008) & 0.121069 (19.1\%) & 592 \\
\midrule

$C_1$ 
& $N=200$ (0.244586) & $N=176$ (0.161373) & 0.083213 (34\%) & 174 \\
& $N=300$ (0.358582) & $N=293$ (0.267289) & 0.091293 (25.5\%) & 285 \\
& $N=500$ (0.405852) & $N=500$ (0.405852) & -- & 497 \\
& $N=700$ (0.635077) & $N=698$ (0.588222) & 0.046855 (7.4\%) & 688 \\
\midrule

$C_2$ 
& $N=200$ (0.244586) & $N=195$ (0.221119) & 0.023467 (9.6\%) & 190 \\
& $N=300$ (0.358582) & $N=293$ (0.277762) & 0.080820 (22.5\%) & 277 \\
& $N=500$ (0.405852) & $N=498$ (0.390607) & 0.015245 (3.8\%) & 493 \\
& $N=700$ (0.635077) & $N=682$ (0.566703) & 0.068374 (10.8\%) & 664 \\
\midrule

$C_3$ 
& $N=200$ (0.244586) & $N=200$ (0.244586) & -- & 198 \\
& $N=300$ (0.358582) & $N=294$ (0.338173) & 0.020409 (5.7\%) & 286 \\
& $N=500$ (0.405852) & $N=471$ (0.253978) & 0.151874 (37.4\%) & 450 \\
& $N=700$ (0.635077) & $N=697$ (0.570629) & 0.064448 (10.1\%) & 677 \\
\bottomrule
\end{tabular*}
\end{table*}

\begin{figure*}[!h]
\centering
\includegraphics[width=0.9\textwidth, height=0.55\textwidth ]{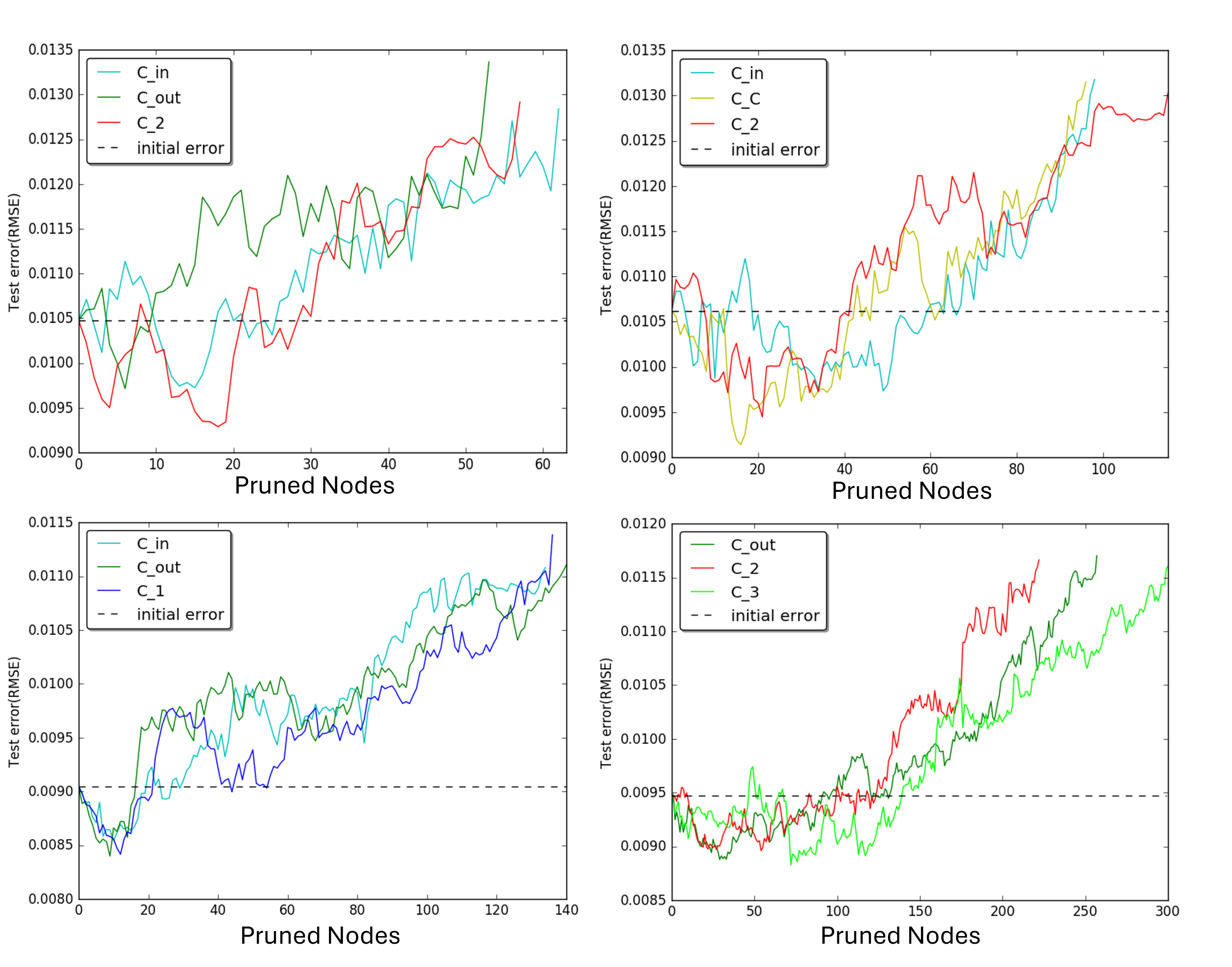}
\caption{Test RMSE versus pruned nodes for Mackey-Glass prediction across reservoir sizes $N=200$ (top left), $N=300$ (top right), $N=500$ (bottom left), and $N=700$ (bottom right). The curves show different centrality measures, and the dashed line indicates the initial error. In general, moderate pruning improves performance before it degrades.}
\label{fig:mackey_pruning}
\end{figure*}

\begin{figure*}[!h]
\centering
\includegraphics[width=0.9\textwidth, height=0.55\textwidth]{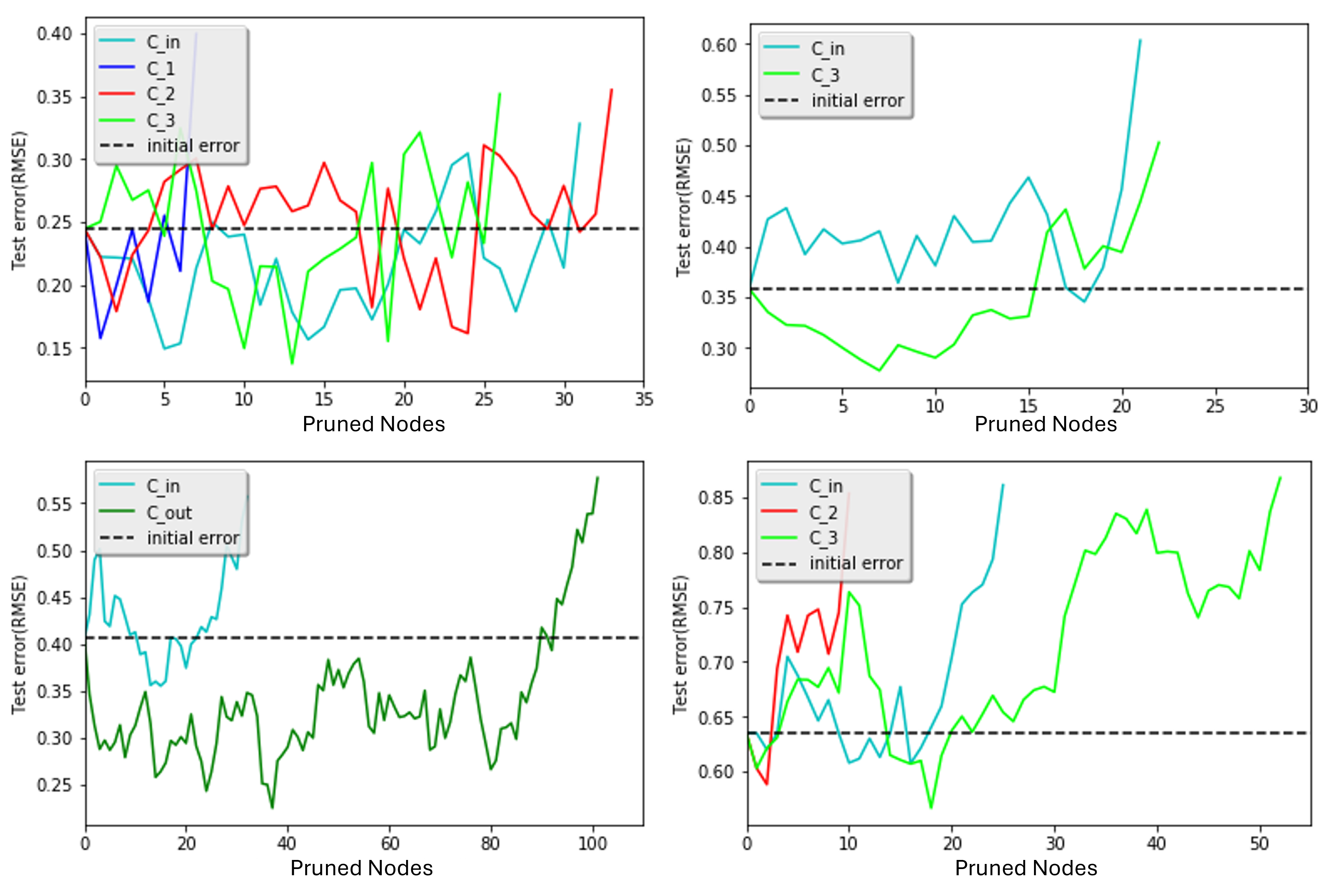}
\caption{Performance of centrality-based pruning on electricity load forecasting, shown as test RMSE versus pruned nodes for reservoir sizes $N=200$ (top left), $N=300$ (top right), $N=500$ (bottom left), and $N=700$ (bottom right). Different centrality measures are compared, and the dashed line shows the initial error.}
\label{fig:electricity_pruning}
\end{figure*}

% \begin{figure*}[!h]
% \centering
% \includegraphics[width=0.9\textwidth, height=0.55\textwidth  ]{./figures/figure2.png}
% \caption{Performance of centrality-based pruning on electricity load forecasting, shown as test RMSE versus pruned nodes for reservoir sizes $N=200$ (top left), $N=300$ (top right), $N=500$ (bottom left), and $N=700$ (bottom right). Different centrality measures are compared, and the dashed line shows the initial error.}
% \end{figure*}

%=====================================

\section{Experimental Setup}

To evaluate the proposed pruning approach, we use the normalized Root Mean Square Error (RMSE), defined as
\begin{equation}
RMSE =
\sqrt{
\frac{1}{N_r\sigma^2}
\sum_{i=1}^{N_r}
(\hat{o}_i-o_i)^2
},
\end{equation}
where $N_r$ denotes the number of prediction points and $\sigma^2$ is the variance of the target signal.

In all experiments, the spectral radius of the reservoir weight matrix is maintained below one to satisfy the echo state property. The sparsity and recurrent structure of the reservoir are also retained after pruning, so that the reduced model remains within the ESN framework.

\subsection{Mackey-Glass Time Series}

The Mackey-Glass chaotic time-series is a standard benchmark for evaluating nonlinear prediction models \citep{mackey1977oscillation, jaeger2004harnessing}. It is defined by the delay differential equation

\begin{equation}
\frac{\partial o(t)}{\partial t}
=
\frac{0.2\,o(t-\alpha)}{1+o(t-\alpha)^{10}}
-0.1\,o(t),
\end{equation}

where $\alpha$ represents the delay parameter. In our experiments, we use $\alpha=17$, which produces highly nonlinear temporal dynamics.

The dataset consists of 10,000 generated time-series samples. The first 10\% of the data is used for initialization, 80\% for training, and the remaining 20\% is equally split into validation and testing sets. For each configuration, 84-step- ahead predictions are performed, and the experiment is repeated 50 times to obtain stable RMSE estimates.

\subsection{Electric Load Data}

To evaluate the proposed method in a real-world setting, we use electricity consumption data obtained from AI Hub (\url{https://www.aihub.or.kr/}). These datasets exhibit complex temporal patterns and are widely used for benchmarking forecasting models.

We follow the same data splitting strategy as in the Mackey-Glass experiments, using 10\% of the data for initialization, 80\% for training, and the remaining 20\% equally divided into validation and testing sets. We consider reservoir sizes of $N=200$, $300$, $500$, and $700$ to analyze the effect of centrality-based pruning.

\subsection{Evaluation Protocol}

We evaluate the proposed method relative to the standard unpruned ESN, which serves as the primary baseline. This controlled comparison isolates the effect of centrality-based pruning on reservoir structure, efficiency, and prediction performance.

Across different reservoir sizes, the pruning process removes approximately 25\% to 40\% of the nodes, depending on the configuration. This enables us to analyze the trade-off between model complexity and predictive performance under varying pruning levels.

All experiments were conducted on a CPU-based environment without GPU acceleration. Since ESN training primarily involves solving a linear regression problem for the readout layer, the computational requirements remain relatively low.

While comparisons with other forecasting models are important, the focus of this work is on structural optimization within the ESN framework. Therefore, we first validate the effectiveness of the proposed approach in improving the baseline ESN. Extensions to comparisons with other state-of-the-art models are left for future work.

%===============================
\section{Results}

The experimental results demonstrate that centrality-based pruning can effectively reduce reservoir size while maintaining or improving prediction accuracy across different tasks and reservoir sizes (see Tables~\ref{tab:mackey_all_models} and \ref{tab:electricity_all_models}).

For the Mackey-Glass task, pruning consistently reduces RMSE as the number of reservoir nodes decreases. In particular, the $C_2$ measure achieves the most consistent improvements across different reservoir sizes, suggesting that considering both incoming and outgoing connections provides a better representation of node influence in chaotic dynamics. Since Mackey-Glass involves complex temporal dependencies, nodes that contribute to both information reception and propagation appear to be more important. As a result, removing nodes with low $C_2$ values helps eliminate redundant structure without degrading chaotic prediction performance.

For electric load forecasting, a different behavior is observed. In this case, $C_{out}$ often achieves the largest improvements, especially for larger reservoirs. This indicates that outgoing connections play a more important role in this task, where the propagation of information through the reservoir has a stronger impact on prediction performance. Removing nodes with weak outgoing influence helps reduce noise and improves generalization.

Across both tasks, moderate pruning generally leads to improved performance, while excessive pruning eventually degrades accuracy. This trend is clearly visible in the performance curves in Fig.~\ref{fig:mackey_pruning} and Fig.~\ref{fig:electricity_pruning}, where RMSE initially decreases and then increases as more nodes are removed. This behavior suggests that some nodes are redundant, but excessive pruning may remove weakly connected nodes that still contribute indirectly to long-term dynamics.

Additionally, separating positive and negative weights provides a more accurate characterization of node influence. Measures such as $C_1$ and $C_2$, which explicitly account for this separation, consistently outperform simple degree-based metrics in several settings (see Tables~\ref{tab:mackey_all_models} and \ref{tab:electricity_all_models}). In contrast, the $C_3$ measure, which only considers the total magnitude of connections, does not consistently improve performance. This suggests that overall connectivity alone is not sufficient to capture the functional importance of nodes. Since $C_3$ ignores the balance between positive and negative weights, it may retain nodes that are strongly connected but do not contribute meaningfully to the reservoir dynamics.

Overall, graph-based structural analysis provides a simple and effective approach for optimizing ESN reservoirs. The effectiveness of each centrality measure depends on the task and the underlying dynamics of the system.

%------------------------------------------------

\section{Conclusion}

This work demonstrates that graph centrality provides a simple and effective criterion for identifying redundant nodes in Echo State Network reservoirs. By representing the reservoir as a weighted directed graph, the proposed approach ranks nodes according to their structural importance and removes low-centrality units to obtain a more compact model. The results show that substantial reservoir reduction is possible without sacrificing prediction accuracy, and in several cases pruning also improves generalization by eliminating weak or redundant dynamics. These findings suggest that graph-based structural analysis is a promising direction for designing more efficient reservoir computing models, with future extensions including spectral centrality measures, adaptive pruning strategies, and applications to broader recurrent architectures.

%------------------------------------------------

% \bibliography{example_paper}

\begin{thebibliography}{36}
\providecommand{\natexlab}[1]{#1}
\providecommand{\url}[1]{\texttt{#1}}
\expandafter\ifx\csname urlstyle\endcsname\relax
  \providecommand{\doi}[1]{doi: #1}\else
  \providecommand{\doi}{doi: \begingroup \urlstyle{rm}\Url}\fi

\bibitem[Ayoub et~al.(2022)Ayoub, Lotfi, and Hammouch]{ayoub2022link}
Ayoub, J., Lotfi, D., and Hammouch, A.
\newblock Link prediction using betweenness centrality and graph neural networks.
\newblock \emph{Social Network Analysis and Mining}, 13\penalty0 (1):\penalty0 5, 2022.

\bibitem[Bala et~al.(2018)Bala, Ismail, Ibrahim, and Sait]{bala2018applications}
Bala, A., Ismail, I., Ibrahim, R., and Sait, S.~M.
\newblock Applications of metaheuristics in reservoir computing techniques: a review.
\newblock \emph{Ieee Access}, 6:\penalty0 58012--58029, 2018.

\bibitem[Battaglia et~al.(2018)Battaglia, Hamrick, Bapst, Sanchez-Gonzalez, Zambaldi, Malinowski, Tacchetti, Raposo, Santoro, Faulkner, et~al.]{battaglia2018relational}
Battaglia, P.~W., Hamrick, J.~B., Bapst, V., Sanchez-Gonzalez, A., Zambaldi, V., Malinowski, M., Tacchetti, A., Raposo, D., Santoro, A., Faulkner, R., et~al.
\newblock Relational inductive biases, deep learning, and graph networks.
\newblock \emph{arXiv preprint arXiv:1806.01261}, 2018.

\bibitem[Bianchi et~al.(2016)Bianchi, Livi, and Alippi]{bianchi2016investigating}
Bianchi, F.~M., Livi, L., and Alippi, C.
\newblock Investigating echo-state networks dynamics by means of recurrence analysis.
\newblock \emph{IEEE transactions on neural networks and learning systems}, 29\penalty0 (2):\penalty0 427--439, 2016.

\bibitem[Bloch et~al.(2023)Bloch, Jackson, and Tebaldi]{bloch2023centrality}
Bloch, F., Jackson, M.~O., and Tebaldi, P.
\newblock Centrality measures in networks.
\newblock \emph{Social Choice and Welfare}, 61\penalty0 (2):\penalty0 413--453, 2023.

\bibitem[Chen et~al.(2016)Chen, Saad, and Yin]{chen2016echo}
Chen, M., Saad, W., and Yin, C.
\newblock Echo state networks for self-organizing resource allocation in lte-u with uplink--downlink decoupling.
\newblock \emph{IEEE Transactions on Wireless Communications}, 16\penalty0 (1):\penalty0 3--16, 2016.

\bibitem[Chowdhury et~al.(2025)Chowdhury, Katlariwala, and Kashyap]{chowdhury2025effective}
Chowdhury, S., Katlariwala, S.~B., and Kashyap, D.
\newblock Effective fine-tuning with eigenvector centrality based pruning.
\newblock In \emph{International Symposium on Visual Computing}, pp.\  3--15. Springer, 2025.

\bibitem[Estrada(2012)]{estrada2012structure}
Estrada, E.
\newblock \emph{The structure of complex networks: theory and applications}.
\newblock American Chemical Society, 2012.

\bibitem[Estrada \& Rodriguez-Velazquez(2005)Estrada and Rodriguez-Velazquez]{estrada2005subgraph}
Estrada, E. and Rodriguez-Velazquez, J.~A.
\newblock Subgraph centrality in complex networks.
\newblock \emph{Physical Review E—Statistical, Nonlinear, and Soft Matter Physics}, 71\penalty0 (5):\penalty0 056103, 2005.

\bibitem[Fan et~al.(2019)Fan, Zeng, Ding, Chen, Sun, and Liu]{fan2019learning}
Fan, C., Zeng, L., Ding, Y., Chen, M., Sun, Y., and Liu, Z.
\newblock Learning to identify high betweenness centrality nodes from scratch: A novel graph neural network approach.
\newblock In \emph{Proceedings of the 28th ACM international conference on information and knowledge management}, pp.\  559--568, 2019.

\bibitem[Freeman(1978)]{freeman1978centrality}
Freeman, L.~C.
\newblock Centrality in social networks conceptual clarification.
\newblock \emph{Social networks}, 1\penalty0 (3):\penalty0 215--239, 1978.

\bibitem[Gallicchio et~al.(2017)Gallicchio, Micheli, and Pedrelli]{gallicchio2017deep}
Gallicchio, C., Micheli, A., and Pedrelli, L.
\newblock Deep reservoir computing: A critical experimental analysis.
\newblock \emph{Neurocomputing}, 268:\penalty0 87--99, 2017.

\bibitem[Gallicchio et~al.(2018)Gallicchio, Micheli, and Pedrelli]{gallicchio2018design}
Gallicchio, C., Micheli, A., and Pedrelli, L.
\newblock Design of deep echo state networks.
\newblock \emph{Neural Networks}, 108:\penalty0 33--47, 2018.

\bibitem[Gonbadi et~al.(2025)Gonbadi, Rostami, Sahafizadeh, Rostami, Nejad, and Shirzadi]{gonbadi2025input}
Gonbadi, L., Rostami, H., Sahafizadeh, E., Rostami, S., Nejad, M.~M., and Shirzadi, A.
\newblock Input driven optimization of echo state network parameters for prediction on chaotic time series.
\newblock \emph{Scientific Reports}, 15\penalty0 (1):\penalty0 33005, 2025.

\bibitem[Huang et~al.(2023)Huang, Wang, Yang, and Li]{huang2023semi}
Huang, J., Wang, F., Yang, X., and Li, Q.
\newblock Semi-supervised echo state network with partial correlation pruning for time-series variables prediction in industrial processes.
\newblock \emph{Measurement Science and Technology}, 34\penalty0 (9):\penalty0 095106, 2023.

\bibitem[Jaeger(2001)]{jaeger2001short}
Jaeger, H.
\newblock Short term memory in echo state networks.
\newblock 2001.

\bibitem[Jaeger(2007)]{jaeger2007echo}
Jaeger, H.
\newblock Echo state network.
\newblock \emph{scholarpedia}, 2\penalty0 (9):\penalty0 2330, 2007.

\bibitem[Jaeger \& Haas(2004)Jaeger and Haas]{jaeger2004harnessing}
Jaeger, H. and Haas, H.
\newblock Harnessing nonlinearity: Predicting chaotic systems and saving energy in wireless communication.
\newblock \emph{science}, 304\penalty0 (5667):\penalty0 78--80, 2004.

\bibitem[Kitsak et~al.(2010)Kitsak, Gallos, Havlin, Liljeros, Muchnik, Stanley, and Makse]{kitsak2010identification}
Kitsak, M., Gallos, L.~K., Havlin, S., Liljeros, F., Muchnik, L., Stanley, H.~E., and Makse, H.~A.
\newblock Identification of influential spreaders in complex networks.
\newblock \emph{Nature physics}, 6\penalty0 (11):\penalty0 888--893, 2010.

\bibitem[Li et~al.(2019)Li, Zhu, and Sun]{li2019deep}
Li, L., Zhu, J., and Sun, M.-T.
\newblock Deep learning based method for pruning deep neural networks.
\newblock In \emph{2019 IEEE International Conference on Multimedia \& Expo Workshops (ICMEW)}, pp.\  312--317. IEEE, 2019.

\bibitem[Liu et~al.(2022)Liu, Bai, Jin, Wang, Su, and Kong]{liu2022broad}
Liu, W., Bai, Y., Jin, X., Wang, X., Su, T., and Kong, J.
\newblock Broad echo state network with reservoir pruning for nonstationary time series prediction.
\newblock \emph{Computational Intelligence and Neuroscience}, 2022\penalty0 (1):\penalty0 3672905, 2022.

\bibitem[Luko{\v{s}}evi{\v{c}}ius et~al.(2012)Luko{\v{s}}evi{\v{c}}ius, Jaeger, and Schrauwen]{lukovsevivcius2012reservoir}
Luko{\v{s}}evi{\v{c}}ius, M., Jaeger, H., and Schrauwen, B.
\newblock Reservoir computing trends.
\newblock \emph{KI-K{\"u}nstliche Intelligenz}, 26\penalty0 (4):\penalty0 365--371, 2012.

\bibitem[Maass et~al.(2002)Maass, Natschl{\"a}ger, and Markram]{maass2002real}
Maass, W., Natschl{\"a}ger, T., and Markram, H.
\newblock Real-time computing without stable states: A new framework for neural computation based on perturbations.
\newblock \emph{Neural computation}, 14\penalty0 (11):\penalty0 2531--2560, 2002.

\bibitem[Mackey \& Glass(1977)Mackey and Glass]{mackey1977oscillation}
Mackey, M.~C. and Glass, L.
\newblock Oscillation and chaos in physiological control systems.
\newblock \emph{Science}, 197\penalty0 (4300):\penalty0 287--289, 1977.

\bibitem[Nakajima \& Fischer(2021)Nakajima and Fischer]{nakajima2021reservoir}
Nakajima, K. and Fischer, I.
\newblock \emph{Reservoir computing}.
\newblock Springer, 2021.

\bibitem[Newman(2018)]{newman2018networks}
Newman, M.
\newblock \emph{Networks}.
\newblock Oxford university press, 2018.

\bibitem[Ozturk et~al.(2007)Ozturk, Xu, and Principe]{ozturk2007analysis}
Ozturk, M.~C., Xu, D., and Principe, J.~C.
\newblock Analysis and design of echo state networks.
\newblock \emph{Neural computation}, 19\penalty0 (1):\penalty0 111--138, 2007.

\bibitem[Pathak et~al.(2018)Pathak, Hunt, Girvan, Lu, and Ott]{pathak2018model}
Pathak, J., Hunt, B., Girvan, M., Lu, Z., and Ott, E.
\newblock Model-free prediction of large spatiotemporally chaotic systems from data: A reservoir computing approach.
\newblock \emph{Physical review letters}, 120\penalty0 (2):\penalty0 024102, 2018.

\bibitem[Perra \& Fortunato(2008)Perra and Fortunato]{perra2008spectral}
Perra, N. and Fortunato, S.
\newblock Spectral centrality measures in complex networks.
\newblock \emph{Physical Review E—Statistical, Nonlinear, and Soft Matter Physics}, 78\penalty0 (3):\penalty0 036107, 2008.

\bibitem[Rodan \& Tino(2010)Rodan and Tino]{rodan2010minimum}
Rodan, A. and Tino, P.
\newblock Minimum complexity echo state network.
\newblock \emph{IEEE transactions on neural networks}, 22\penalty0 (1):\penalty0 131--144, 2010.

\bibitem[Scardapane et~al.(2014)Scardapane, Nocco, Comminiello, Scarpiniti, and Uncini]{scardapane2014effective}
Scardapane, S., Nocco, G., Comminiello, D., Scarpiniti, M., and Uncini, A.
\newblock An effective criterion for pruning reservoir's connections in echo state networks.
\newblock In \emph{2014 International joint conference on neural networks (IJCNN)}, pp.\  1205--1212. IEEE, 2014.

\bibitem[Sima et~al.(2025)Sima, Bao, Zhang, He, and Lai]{sima2025enhancing}
Sima, Q., Bao, Y., Zhang, X., He, K., and Lai, X.
\newblock Enhancing echo state network with reservoir state selection for time series forecasting.
\newblock \emph{Neurocomputing}, pp.\  131283, 2025.

\bibitem[Sun et~al.(2022)Sun, Song, Cai, Zhang, Hong, and Li]{sun2022systematic}
Sun, C., Song, M., Cai, D., Zhang, B., Hong, S., and Li, H.
\newblock A systematic review of echo state networks from design to application.
\newblock \emph{IEEE Transactions on Artificial Intelligence}, 5\penalty0 (1):\penalty0 23--37, 2022.

\bibitem[Tanaka et~al.(2019)Tanaka, Yamane, H{\'e}roux, Nakane, Kanazawa, Takeda, Numata, Nakano, and Hirose]{tanaka2019recent}
Tanaka, G., Yamane, T., H{\'e}roux, J.~B., Nakane, R., Kanazawa, N., Takeda, S., Numata, H., Nakano, D., and Hirose, A.
\newblock Recent advances in physical reservoir computing: A review.
\newblock \emph{Neural Networks}, 115:\penalty0 100--123, 2019.

\bibitem[Valente et~al.(2008)Valente, Coronges, Lakon, and Costenbader]{valente2008correlated}
Valente, T.~W., Coronges, K., Lakon, C., and Costenbader, E.
\newblock How correlated are network centrality measures?
\newblock \emph{Connections (Toronto, Ont.)}, 28\penalty0 (1):\penalty0 16, 2008.

\bibitem[Wang et~al.(2024)Wang, Lun, Li, and Lu]{wang2024multi}
Wang, B., Lun, S., Li, M., and Lu, X.
\newblock Multi-reservoir echo state network with five-elements cycle.
\newblock \emph{Information Sciences}, 661:\penalty0 120166, 2024.

\end{thebibliography}
% \bibliographystyle{icml2026}
\bibliographystyle{icml2026}

%%%%%%%%%%%%%%%%%%%%%%%%%%%%%%%%%%%%%%%%%%%%%%%%%%%%%%%%%%%%%%%%%%%%%%%%%%%%%%%
%%%%%%%%%%%%%%%%%%%%%%%%%%%%%%%%%%%%%%%%%%%%%%%%%%%%%%%%%%%%%%%%%%%%%%%%%%%%%%%

\end{document}